# Probabilistic Saliency Estimation

Caglar Aytekin, Alexandros Iosifidis and Moncef Gabbouj

**Abstract— In this paper, we model the salient object detection problem under a probabilistic framework encoding the boundary connectivity saliency cue and smoothness constraints in an optimization problem. We show that this problem has a closed form global optimum which estimates the salient object. We further show that along with the probabilistic framework, the proposed method also enjoys a wide range of interpretations, i.e. graph cut, diffusion maps and one-class classification. With an analysis according to these interpretations, we also find that our proposed method provides approximations to the global optimum to another criterion that integrates local/global contrast and large area saliency cues. The proposed approach achieves mostly leading performance compared to the state-of-the-art algorithms over a large set of salient object detection datasets including around 17k images for several evaluation metrics. Furthermore, the computational complexity of the proposed method is favorable/comparable to many state-of-the-art techniques.**

*Index Terms*—Saliency, salient object detection, spectral graph cut, diffusion maps, probabilistic model, one-class classification.

## I. INTRODUCTION

Salient object detection is a computer vision research topic with growing interest over the last decade. The goal is to highlight the visually interesting regions in a given scene. The problem of saliency detection has been motivated by related work in neuroscience; humans select important visual information based on attention mechanisms in the brain [1]. Given this motivation, earlier works on saliency detection concentrate more on predicting sparse human eye-gaze points that are detected by eye-trackers [2]. Accordingly, most of the research on this track is based on biologically inspired algorithms which try to imitate the dynamics of the human attention mechanism [2]-[9]. During the last few years another related track emerged where the goal is to segment out *salient objects* [10]-[20], instead of predicting some sparse eye-fixations. Both research tracks produce saliency maps that are useful for tasks such as video surveillance [21], compression [22], image manipulation [23], automatic image cropping [24], foreground detection [25], and coding [26], to name a few. However, the output of salient object detector techniques is more useful, when compared to eye fixation predictions, for higher level computer vision and pattern recognition

C. Aytekin, A. Iosifidis and M. Gabbouj are with the Department of Signal Processing, Tampere University of Technology, P.O.Box 553, FI-33101, Tampere, Finland.

tasks such as tracking [27], object region proposals [29] and object recognition [28].

In this paper, we focus on the salient object detection task. Since ultimately we consider salient object detection as a pre-processing block for higher level tasks as mentioned above, fast and generic methods would be preferred. Therefore, in this work we focus on unsupervised salient object detection. While supervised approaches, such as those in [45] and [36], have the potential of finding more accurate results, their performance depends on the training process followed and the data that has been exploited for training. Recent works have also indicated that unsupervised saliency detection approaches can compete (or even outperform) supervised methods [43].

Unsupervised salient object detection methods can be categorized based on the saliency cues they use. Commonly exploited cues include local and global contrast, boundary connectivity, shape and location cues. The local contrast cue is based on the assumption that the salient object is in contrast with its immediate surroundings [11], [12], [15], [20]. A spectral foreground detection method was proposed in [20] which optimizes a criterion involving the minimization of cut-value, which is equivalent to the maximization of the local contrast. A region contrast based method was proposed in [11], which computes a salient region as a weighted sum of local contrasts with its surroundings. The global contrast cue is similarly defined as by assuming that the salient object is in high contrast with its surrounding. A histogram-based method enhancing regions with global contrast to the rest of the image was proposed in [11]. The boundary connectivity prior is one of the most widely used cues and is based on the assumption that most of the image boundaries will not contain parts of the salient object [10], [15], [16], [20], [39]. There are many ways to use this prior. For example in [20], the boundary pixels were strongly assigned to background in a belief vector, which was a part of the optimization problem used to extract salient regions. A robust background assignment on image boundaries was conducted in [15] based on a handcrafted measure that differentiates foreground and background regions touching the boundary. The boundary regions were used as background templates in [10] to re-construct the image in a sparse and dense way. Salient regions were defined as the ones that have high reconstruction errors [10]. In [16], boundary regions were defined as absorbing nodes in a Markov chain model. The saliency values of each region were evaluated based on the absorption time of the regions by the boundary nodes. Similar to the boundary connectivity cue, the center prior also assumes that salient objects are less likely to touch the image boundaries. However, it makes a stronger assumption that the object is mostly located in the center of the image. Although this is not a very reliable cue, it was used in [12] as a weighting coefficient on local contrast based saliency maps.

Besides the saliency cues, there are other supplementary processes that are widely used, such as using multiple resolutions [10], [12], [20] and smoothness constraints on saliency maps within similar regions [15], [20].

The salient object detection methods can also be categorized according to their interpretation of saliency. One of the most popular interpretations is modelling saliency as a diffusion process [14], [16], [39]. These methods assume a preliminary information on saliency and designs several diffusion matrices to propagate this initial information to the whole salient region. Another interpretation of saliency is a graph-cut based one, where the salient object is considered as a foreground partition which constitutes a large area and a high contrast with the background partition [19],[20],[29]. The concepts behind these interpretations, i.e. diffusion maps and graph partitioning were shown to be related in an earlier work [38].

For an extensive survey on the recent state-of-the-art salient object detection algorithms, readers are encouraged to see [37]. Furthermore, an extensive benchmark comparing the performance of the state-of-the-art in salient object detection is also provided in [43].

In this paper, we model the salient object detection problem by following a probabilistic approach. We propose an unsupervised salient object detection method (which we call Probabilistic Saliency Estimation – PSE) that jointly optimizes saliency cues such as boundary connectivity and smoothness constraints. The solution of the proposed optimization problem is shown to have a closed-form. Moreover, we show that, along with the proposed probabilistic framework, this solution can also be interpreted from several perspectives, including spectral graph cut, diffusion and one-class classification approaches, all leading to the exact same solution. By further exploiting these links, we show that some diffusion based [14], [16], [39] and spectral graph based [20] salient object detection methods can also be cast into the proposed probabilistic framework, which provides us a theoretical platform for comparison of our method with them. Based on the above analysis provided in this paper, we also show that the proposed PSE method differentiates from our previous methods [19],[20] from three aspects. First, it introduces a new formulation for saliency detection in a probabilistic framework. Second, by exploiting the graph-cut interpretation of PSE, we show that it also provides another approximation of the optimization criterion exploited in [19],[20] with looser constraints. Finally, we show that [19],[20] are suboptimal when they are investigated in the proposed probabilistic framework.

The main contributions of the paper can be listed as follows:

- A novel probabilistic approach to the unsupervised salient object detection problem is proposed. Following this approach, we formulate a novel optimization problem and we show that it has a global optimal solution (PSE), leading to state-of-the-art performance for unsupervised salient object detection. (Section II)

- Based on the links observed between the proposed solution and those of diffusion based salient object detection methods, we interpret PSE as a diffusion based method. (Section II.B)

- Based on the analysis of diffusion interpretation of PSE, we show that its solution involves terms related to graph cut. Therefore, PSE can also be interpreted as a solution to the graph-based cut problem. In fact, we show that PSE can be derived directly from the original graph based cut problem. (Section II.C)

- We show that some diffusion based and graph based methods can be cast into our proposed probabilistic framework under

some assumptions. (Appendices A,B)

- Finally, we show that the saliency detection problem can be interpreted as a one-class classification problem, whose solution is also given by PSE. (Section II.D)

The rest of the manuscript is organized as follows. In Section II, we introduce the proposed PSE method and explore its connections and differences to a wide range of salient object detection methods, namely the diffusion, the graph-based, and the one-class classification approaches. In Section III, we present extensive experiments conducted on publically available datasets and compare the performance of the proposed method with the state-of-the-art. Finally, we conclude the paper in Section IV.

## II. PROBABILISTIC SALIENCY ESTIMATION

We formulate the salient object detection problem as that of the estimation of the probability mass function (PMF) $P(\mathbf{x})$ of a random variable $\mathbf{x}$, where a region/pixel $x_i$ in an image is a possible outcome of the event. Since the summation over the PMF is constraint to be equal to 1, such a formulation allows us to set the total attention given to the scene as fixed which is analogous to the limited processing capability of the human brain. Accordingly, a study on the mechanisms of visual attention in the human cortex [1] states that: "*A typical scene contains many different objects that, because of the limited processing capacity of the visual system, compete for neural representation*". The probabilistic framework allows us to model such competition by defining an event as drawing distinct regions from the image, where $P(\mathbf{x})$ defines the probability mass function of this event such that $P(\mathbf{x} = x_i)$ is high if the region $x_i$ is salient. Within this framework, we wish $P(\mathbf{x})$ to satisfy some properties related to the widely used saliency assumptions. For example, we expect similar probabilities for the image regions that are similar in a feature space, such as a color space. This is in accordance with the smoothness constraints for saliency as explained above. Furthermore, we wish to design $P(\mathbf{x})$ in such a way that any prior information about the PMF can be considered. For example, in this way one can make use of the widely accepted assumption that image boundaries are more likely to belong to the non-salient region. Therefore, the estimation of $P(\mathbf{x})$ can be formulated as an optimization problem that involves two terms: one that enforces similar regions in a given feature space to have similar probabilities and another that encodes any prior information about the PMF. We express this joint optimization as follows:

$$\underset{P(\mathbf{x})}{argmin} \left( \sum_i \big( P(\mathbf{x} = x_i) \big)^2 v_i + \frac{1}{2} \sum_{i,j} \big( P(\mathbf{x} = x_i) - P(\mathbf{x} = x_j) \big)^2 w_{i,j} \right) \qquad (1)$$

$$s.t. \quad \sum_i P(\mathbf{x} = x_i) = 1.$$

In (1), the first term suppresses the probability of a region $x_i$ if there is prior information that this region belongs to a non-salient region. This belief is encoded as a positive $v_i$ value. The second term in (1) forces two regions $x_i$ and $x_j$ to have similar probabilities if their similarity $w_{i,j}$ in a feature space is high. Here, we avoid using an additional parameter for the regularization of the two terms in this optimization problem, since this regularization can be encoded by scaling the values of $\boldsymbol{v}$. For any symmetric similarity function, i.e. when $w_{i,j} = w_{j,i}$, one can rewrite (1) as follows:

$$\underset{P(\mathbf{x})}{argmin} \left( \sum_i \left( P(\mathbf{x} = x_i) \right)^2 v_i + \left( \sum_{i,j} \left( \left( P(\mathbf{x} = x_i) \right)^2 - P(\mathbf{x} = x_i) P(\mathbf{x} = x_j) \right) w_{i,j} \right) \right) \tag{2}$$

$$s.t. \quad \sum_i P(\mathbf{x} = x_i) = 1.$$

In order to simplify the notation, let us introduce a vector $\boldsymbol{p}$ having elements $p_i = P(\mathbf{x} = x_i)$. Let us also define a diagonal matrix $\mathbf{V}$ having elements $[\mathbf{V}]_{ii} = v_i$. Then, we can rewrite (2) in matrix form as follows:

$$p^* = \underset{p}{argmin} \left( p^T H p \right)$$

$$s.t. \quad p^T 1 = 1, \tag{3}$$

In (3), $\mathbf{H} = \mathbf{D} - \mathbf{W} + \mathbf{V}$, where $\mathbf{W}$ is a matrix containing the similarity values $w_{i,j}$ and $\mathbf{D}$ is a diagonal matrix whose elements are obtained as $[\mathbf{D}]_{ii} = \sum_j w_{i,j}$.

In order to obtain the optimal solution of the optimization problem in (3), we calculate the saddle points of the Lagrangian given by:

$$\mathcal{L}(\boldsymbol{p}, \gamma) = \left( p^T H p \right) - \gamma \left( p^T 1 - 1 \right). \tag{4}$$

Partial derivative of $\mathcal{L}(\boldsymbol{p}, \gamma)$ with respect to $\boldsymbol{p}$ is zero at the saddle points. We can formulate this as follows:

$$\nabla \mathcal{L}(\boldsymbol{p}, \gamma) |_{p^*} = \left( 2 \mathrm{H} p - \gamma 1 \right) = 0. \tag{5}$$

By solving (5), we obtain a single saddle point as follows:

$$p^* = \frac{1}{1^T \mathrm{H}^{-1} 1} \mathrm{H}^{-1} 1 = \mathbf{q}(\mathrm{H}) \, \mathrm{H}^{-1} 1. \tag{6}$$

The normalization term $q(H) = \frac{1}{1^T H^{-1} 1}$ in (6) follows from the constraint $p^T 1 = 1$. Next, we perform the second derivative test to validate that the saddle point (6) is indeed the global minimum of the optimization problem in (3):

$$\frac{\partial^2 \mathcal{L}(p, \gamma)}{\partial p^2} = 2H. \tag{7}$$

Following from (7), if $H$ is positive definite, i.e. it has a positive determinant, the second derivative test guarantees that the saddle point (6) is the global minimum of (4). Following the definition of $H$, i.e. $H = D - W + V$ and the symmetric property of $W$, we observe that $D - W$ is a positive semi-definite matrix. Then, any diagonal matrix $V$ with non-negative entries and at least one positive entry guarantees $H$ to be positive definite. In other words, prior information encoded in $V$ should correspond to a non-salient region and we should have this prior for at least one region.

A constraint that should be satisfied is that all the elements of $p^*$ should be non-negative, which follows from the definition of probability. In order to enforce this constraint, one could include an additional constraint in (3). However, positivity of $p^*$ is already guaranteed from the following facts: (i) $H^{-1}$ is a positive definite matrix, hence the normalization factor in (6) is always positive, i.e., $1^T H^{-1} 1 > 0$, (ii) $H$ is a positive definite $M$-matrix [48], hence $H^{-1}$ is a non-negative matrix [48] which yields non-negativity of the numerator of (6). Finally, the normalization term $q(H)$ scales the values of $p^*$, so that $p^{*T} 1 = 1$. Therefore, $p^*$ is guaranteed to be a valid probability vector.

We define the saliency of each region/pixel $x_i$ as proportional to its probability given by the optimal probability mass function $p^*$ in (6). Therefore, to speed up computations, we can simplify the expression of saliency by removing the positive normalization term $q(H)$ in (6), as follows:

$$p^*_{PSE} = H^{-1} 1. \tag{8}$$

As discussed above, the diagonal matrix $V$ should contain all non-negative values and should have at least one non-zero element in order to have a global minimum of the optimization problem in (3). Hence, if there is no prior information about the probability mass function, we can simply set the same penalty to all samples, i.e. we can set $v_i = r$, for all regions i. This can be considered as a smoothness term for the problem and $r$ can have any non-zero value. As the value of $r$ decreases, the penalty with respect to region similarities is enhanced, whereas a high value of $r$ results in a smoothed map, where all probabilities are similar regardless of the region similarities. On the other hand, if there is any prior information or assumption about non-saliency for some regions, this can be imposed as a positive value in the corresponding elements of $V$ and the remaining elements can be set to zero, meaning that we have no information about them. For example, it is a widely accepted assumption that the image boundaries most likely correspond to background regions [19]. Using such an assumption, we can simply set the diagonal elements of $\mathbf{V}$ that correspond to image boundaries high and leave the rest as zero.

Next, we make an interesting observation from the closed form solution in (8) and note its similarity to the solution of the diffusion based salient object detection methods. We then discuss this similarity and show that PSE can be interpreted as a diffusion based method.

### A. Analysis of Diffusion Based Methods

The saliency detection problem can alternatively be cast as a diffusion problem. A general representation of such algorithms is provided in [39]. This representation is formulated as follows:

$$\mathbf{y}^A = \mathbf{A}^{-1}\mathbf{s}. \tag{9}$$

The saliency map $\mathbf{y}^A$ is calculated by multiplying a seed vector $\mathbf{s}$ with a diffusion matrix $\mathbf{A}^{-1}$. The seed vector is a prior confidence map for saliency. Various diffusion matrices have been used, such as $(\mathbf{D} - \mathbf{W})^{-1}$ in [14] and $\left(\mathbf{D}^{-1}(\mathbf{D} - \mathbf{W})\right)^{-1}$ in [16].

By comparing (8) and (9), one can state that PSE is a diffusion based method using a diffusion matrix $\mathbf{H}^{-1}$ and a seed vector $\mathbf{1}$. In Appendix A, we investigate why our proposed diffusion matrix and seed vector produce better saliency maps when compared to other diffusion methods, following a probabilistic approach.

### B. PSE as a Diffusion Based Method and Links to Graph-Cut

First, let us review a recent finding about diffusion based methods. In [39], it was shown that in diffusion based methods, elements of saliency map were evaluated by a weighted sum of seed vector $\mathbf{s}$, where weights are defined as similarities of diffusion maps given by:

$$\mathbf{y}_i^A = \sum_{j=1}^{N} s_j \langle \Psi_{A_i}^{t=-1/2}, \Psi_{A_j}^{t=-1/2} \rangle,$$

$$\Psi_{A_i}^{t=-1/2} = \left[ \lambda_{A_1}^{-1/2} u_{A_1}(i), \dots, \lambda_{A_N}^{-1/2} u_{A_N}(i) \right]. \tag{10}$$

In (10), $u_{A_n}$ and $\lambda_{A_n}$ denote the $n^{\text{th}}$ eigenvector and eigenvalue of the diffusion matrix $\mathbf{A}$, respectively. The diffusion map $\Psi_{A_1}^{t}$ was defined in [38] as a coordinate system where the Euclidian distance at this coordinate system gives the distance between the probabilities of the states of a Markov Chain at time $t$.

Following our observation in the previous subsection related to the similarity between (8) and (9), PSE can be defined as a diffusion based method where the diffusion matrix is $\mathbf{H}^{-1}$ and the seed vector is $\mathbf{s} = \mathbf{1}$. Thus, we can rewrite $(\mathbf{10})$ for PSE as follows:

$$\mathbf{y}_i^{\mathbf{H}} = \langle \Psi_{\mathbf{H}_i}^{t=-1/2}, \Psi_{\mathbf{H}_i}^{t=-1/2} \rangle + \sum_{j \neq i}^{N} \langle \Psi_{\mathbf{H}_i}^{t=-1/2}, \Psi_{\mathbf{H}_j}^{t=-1/2} \rangle. \tag{11}$$

By analyzing $(\mathbf{11})$, the saliency value of an element $i$ is obtained by summing the similarities of the diffusion map of $i$ to all diffusion maps at $t = -1/2$, i.e., the similarities of the spectral feature vector of $i$ to all feature vectors. As in $(\mathbf{10})$, **the** $i^{\text{th}}$ spectral feature vector is obtained by concatenating the $i^{\text{th}}$ element of the eigenvectors of $\mathbf{H}$, where each eigenvector is weighted by the inverse square-root of the corresponding eigenvalue as a measure of confidence.

Besides the above interpretation, a striking similarity of $(\mathbf{11})$ to one particular spectral foreground detection problem, namely QCut [19] is revealed. In fact, the first term in $(\mathbf{11})$ corresponds to a weighted sum of QCut's saliency maps resulting from exploiting the entire eigen-spectrum (a multispectral approach). In the original work [19], instead of a multispectral approach, only the eigenvector with the corresponding eigenvalue was exploited. However, a multispectral approach was later proved to produce a pool of high quality salient object proposals [29]. In the first term of $(\mathbf{11})$, each QCut saliency map generated from a given eigenvector was weighted with the inverse of the corresponding eigenvalue, which is meaningful because the eigenvalue defines a cost function related to saliency [19]. Next, let us investigate the meaning of the second term of $(\mathbf{11})$. For this, we have to know the structure of the eigenvectors of $\mathbf{H}$. According to QCut's formulation, the squared magnitudes of these eigenvectors are already good indicators of saliency. Hence, it is reasonable to expect the eigenvectors to have high amplitudes in salient regions and close to zero values in non-salient regions. Due to the phase penalty term in QCut, we also expect the entries of the eigenvector to carry the same sign within the same object (i.e. all entries are either non-negative or non-positive). In the light of this information, the second term in $(\mathbf{11})$ turns out to be an enhancement term on QCut's result. Salient objects found by QCut are enhanced by the contribution of the inner product with other salient nodes. Note that due to our assumption that the signs of eigenvectors within the object are the same, we can safely assume that the inner-product is positive; hence this is a reinforcement on the initial saliency maps. Furthermore, if there are two far-away salient objects (whose entries carry different signs) this term will eventually suppress the object with the smaller area. This is reasonable because larger regions are expected to be more salient than smaller ones. In the extreme case, if the area of one object is too small, the object can be considered as noise, and will hence be suppressed. Here we should note that some methods resort to post-processing in order to suppress small objects.

In order to illustrate the effect of the second term in $(\mathbf{11})$ on the saliency map, let us consider the example images shown in Fig. 1. First, we use the first eigenvector of $\mathbf{H}$ to illustrate the effect of the second term on the saliency map. Next, we use the entire eigen-spectrum, which is simply PSE, to highlight the contribution from other eigenvectors. The difference between the use of the first eigenvector and QCut is also illustrated in order to visualize the effect of the enforcement term in **(11)**. We can observe that all elements having non-zero saliency are enhanced with respect to their initial confidence obtained by QCut. In this sense, we can state that although we do not make use of a seed vector **s**, inherently PSE uses a similar confidence weighting where the confidence is determined by the QCut saliency maps.

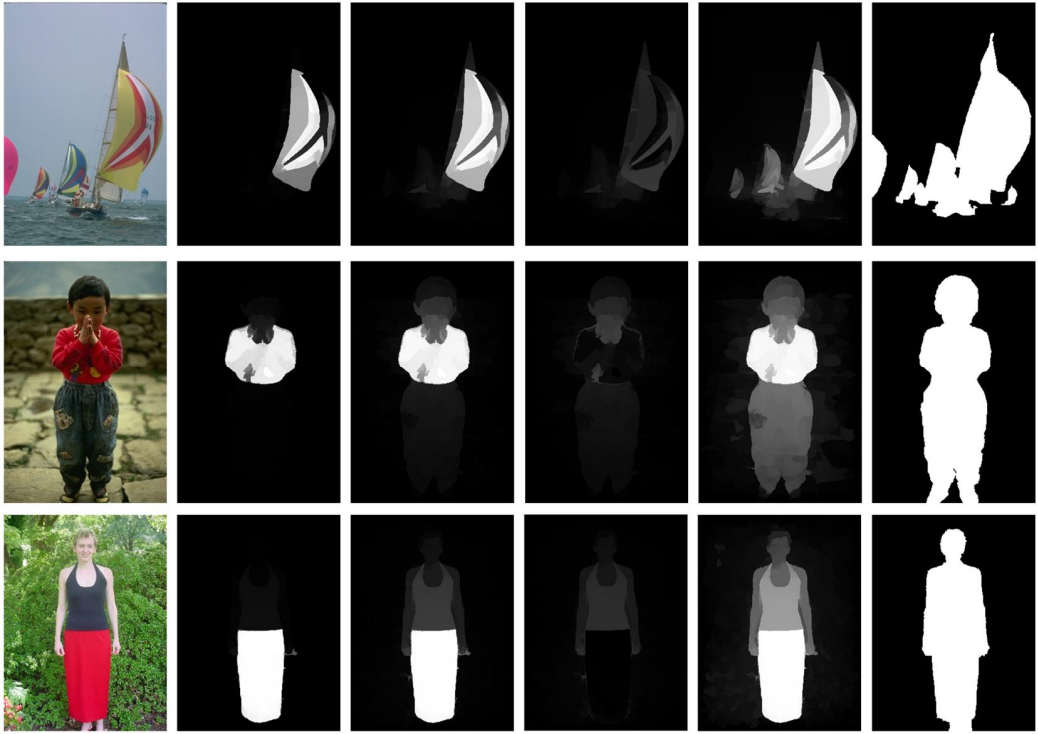

**Fig. 1** (from left to right) An outdoor image, the salient region extracted by QCut [19], PSE with the first eigenvector, contribution of the first eigenvector-based PSE on QCUT, PSE and the ground truth.

By the above analysis, we have observed that the solution of PSE involves enhanced multispectral solutions to an approximation of spectral graph-cut problem [19]. In the next section, we prove that PSE itself is yet another approximation to this specific problem.

*C.   PSE as a Graph-Cut Method*

Following the local and global contrast cues, boundary connectivity and large area cues, a salient object is defined as a collection of pixels (or superpixels) suitably represented on a graph optimizing the following criterion [19].

$$A^* = \underset{A}{\arg\min} \frac{cut(A, \bar{A})}{area(A)} \tag{12}$$

In **(12)**, $\mathbf{A}^*$ is the optimal salient object segment searched over all possible segments of $\mathbf{A}$, **area(A)** is the total number of salient nodes in A and $cut(A, \bar{A}) = \sum_{i \in A, j \in A} w_{i,j}$ is the cost of cutting the edges connecting segment $A$ to the rest of the graph $\bar{\mathbf{A}}$. When edge weights $w_{i,j}$ are defined as affinities, cut-cost can be considered as a measure of contrast. Furthermore, one can add an artificial background node to exploit the boundary cue where every node $i$ is connected to the background node with confidence $\mathbf{v}_i$. One can re-write **(12)** with a binary saliency indicator vector $\mathbf{y}$ as follows:

$$\mathbf{y}_b^* = \operatorname*{argmin}_{\forall i, y_i \in \{0,1\}} \frac{\sum_{i,j} \mathbf{w}_{i,j}(\mathbf{y}_j^2 - \mathbf{y}_i \mathbf{y}_j) + \sum_i \mathbf{v}_i \mathbf{y}_i^2}{\sum_i \mathbf{y}_i}. \tag{13}$$

Eq. **(13)** can be rewritten in matrix form by defining $\mathbf{H} = D - W + V$.

$$\mathbf{y}_b^* = \operatorname*{argmin}_{\forall i, y_i \in \{0,1\}} \frac{\mathbf{y}^{\mathrm{T}} \mathbf{H} \mathbf{y}}{\mathbf{y}^{\mathrm{T}} \mathbf{1}}. \tag{14}$$

The optimization problem in **(14)** is NP-hard. Even if the values of $\mathbf{y}$ are relaxed to have real values, the resulting function is discontinuous at $\mathbf{y} = \mathbf{0}$ and the minimum occurs at $\mathbf{y} \to -\infty$. Obviously, this has no meaning in terms of approximation of the original problem. In [19], the cost function in **(14)** was written in the following form:

$$\mathbf{y}_b^* = \operatorname*{argmin}_{\forall i, y_i \in \{0,1\}} \frac{\sum_{i,j} \mathbf{w}_{i,j}(\mathbf{y}_j - \mathbf{y}_i \mathbf{y}_j) + \sum_i \mathbf{v}_i \mathbf{y}_i}{\sum_i \mathbf{y}_i}, \tag{15}$$

which is equivalent to **(14)** as long as $\forall i, \mathbf{y}_i \in \{0,1\}$. Then, a spectral approximation of this problem was solved by introducing a vector $\mathbf{z}$ that satisfies $\mathbf{y} = \mathbf{z} \circ \mathbf{z}$. Adding a cost of $\sum_{i,j} \mathbf{w}_{i,j}(\mathbf{z}_i^2 \mathbf{z}_j^2 - \mathbf{z}_i \mathbf{z}_j)$ to restrict the solution space of $\mathbf{z}$ by penalizing sign changes, the approximation of **(15)** introduced in [19] can be re-written as follows:

$$\mathbf{y}_{qcut}^* = \operatorname*{argmin}_{\mathbf{y} = \mathbf{z} \circ \mathbf{z}} \frac{\mathbf{z}^{\mathrm{T}} \mathbf{H} \mathbf{z}}{\mathbf{z}^{\mathrm{T}} \mathbf{z}}. \tag{16}$$

Solving the above optimization problem is equivalent to solving the following problem:

$$y^*_{qcut} = \underset{y=z\circ z, \, z^T z = 1}{\textbf{argmin}} \; z^T H z. \tag{17}$$

Eq. **(17)** can be solved via eigen-analysis on **H**, when **z** is relaxed to have real values. This solution has theoretical connections to quantum mechanics and can be interpreted as a probability distribution of a sub-atomic particle's whereabouts in space, and the resulting method was thus called Quantum Cuts (QCut) [19].

Next, we show that PSE provides another approximation to **(12)** using the representation in **(13)** with looser constraints than QCut. We set only the constraint $z^T z = 1$, which can be expressed as $y^T 1 = 1$. We drop all other constraints, i.e., $y = z \circ z$ and the phase-penalty term, and write the new minimization problem that approximates the original problem **(13)** as follows:

$$y^* = \underset{y^T 1 = 1}{\textbf{argmin}} \; y^T H \, y. \tag{18}$$

Notice that **(18)** is equivalent to **(3)**. Hence, we have shown that PSE provides another approximation with looser constraints, when compared to [19], of the graph node cutting problem in **(12)** and $y^* = p^*$.

QCut and PSE approximations of the optimization problem in **(12)** can be formulated as follows:

$$p^* = \frac{1}{c} H^{-1} \mathbf{1}, \qquad c = \mathbf{1}^T H^{-1} \mathbf{1}$$

$$p^*_{Qcut} = z_* \circ z_*, \quad H z_* = \mathbf{E}_m z_*, \tag{19}$$

where $\mathbf{E}_m$ is the minimum eigenvalue of **H.** At first glance, we may observe that both approximations can be related if one would rewrite $p^*$ considering the eigen-decomposition of $H^{-1}$. This was indeed the analysis we made in the previous subsection where we have already shown the benefits of PSE compared to QCut. In Appendix B, we will make another theoretical comparison between QCut and PSE by giving a probabilistic interpretation of QCut and showing the shortcomings of the method in this framework. As the final connection to other approaches, we will show that PSE is also a solution to a one-class classification problem defined for saliency object detection, in the next subsection.

*D.  PSE as a One-Class Classifier*

Let us consider the following problem: Given an image, find a projection vector that will map each image element in a feature space to a fixed value, hence modelling the image with a fixed value. This corresponds to finding a classifier that learns to predict if any feature from a broader set of images fits well to the modelled image or not. Note that this is a one-class classification problem, where we try to map the positive class to a constant value [46]. Let us denote the D-dimensional features of the image as $x_i \in \mathbb{R}^D, i = 1, ..., N$ where $N$ is the number of regions in the image. Now let us represent these features in a space $F$ using a non-linear mapping function $\phi(\cdot)$, such that $x_i \in \mathbb{R}^D \to \phi(x_i) \in F, i = 1, ..., N$. Let us also define the matrix $\Phi = [\phi(x_1), ..., \phi(x_N)]$.

By using this notation, let us denote by $w \in F$ a vector mapping the representations of $x_i$ in $F$ to a fixed value, e.g. the value 1. $w$ can be obtained by solving the following optimization problem:

$$w^* = \underset{w}{arg\,min} \left( \sum_{i=1}^{N} (w^T \phi(x_i) - 1)^2 \right).$$

(20)

In matrix form, we can also rewrite it as follows:

$$w^* = \underset{w}{arg\,min} \| w^T \Phi - 1^T \|_2^2.$$

(21)

Since, in general, $\phi(x_i) \in F$ cannot be obtained, one exploits the Representer Theorem [47] stating that $w$ can be expressed as a linear combination of $\phi(x_i) \in F$, i.e. $w = \Phi y$, where $y \in \mathbb{R}^N$ is a vector encoding the contribution of each image region's features to $w$. Finally, we can rewrite the problem as follows:

$$y^* = \underset{y}{arg\,min} \| y^T \Phi^T \Phi - 1 \|_2^2.$$

(22)

The solution of (22) is given by $y^* = (\Phi^T \Phi)^{-1} 1$. In standard kernel-based one-class classification, the matrix $\Phi^T \Phi$ is replaced by the positive semi-definite kernel matrix [49]. Note that matrix $H$ defined in our probabilistic saliency estimation method is positive definite and hence it can be expressed as $H = \tilde{\Phi}^T \tilde{\Phi}$. Comparing the solution of (22) with (8), we can observe that the proposed probabilistic saliency estimation method can be interpreted as a method solving the above one-class classification problem.

The interpretation given above implies that if an optimal vector that projects each feature of an image to a single value is found, and if this projection vector is represented as a linear combination of features from the image, then the highest contribution to the projection vector will come from the salient regions' features. In other words, when someone tries to describe the content of an image, he/she will focus more on the salient regions of the image. Interestingly, this coincides with the way people describe visual content of an image. For example, the reader may realize when he/she tries to describe an image, e.g. the image in the last row of Fig. 1, that the description will be probably made as follows: "A girl in red skirt and black shirt standing in front of some vegetation next to flowers". Such a description most likely starts with the most salient object/part of the image and less informative regions are mentioned later or not mentioned at all.

At this point, we would like to note that this interpretation is different from the 1-vs-all classification interpretation proposed in [50] where salient object is interpreted as a class and optimal features are pursued that best distinguish this class from others in the image. Our one-class based interpretation differentiates from this approach, as we interpret the whole image as a class (not only the salient region) and we claim that out of all regions in the image, salient regions are the ones that best describe the image content.

Here it should be noted that, based on the above discussion, diffusion methods can also be formulated as classification problems. This is further discussed in Appendix C.

## III. Experimental Results

To evaluate the performance of the proposed PSE method, we conducted extensive experiments in 5 widely used datasets, namely MSRA10K, DUTOMRON, ECSSD, JUDD and SED2. In order to evaluate different aspects of the obtained results, the performance is reported using multiple metrics, i.e. precision-recall curves, receiver operating characteristic curves, mean squared error, maximum F-measure and area under the ROC curve. Before giving a brief description of the datasets and the evaluation metrics, we first provide a few important implementation details.

### A. Experimental set-up

The graph construction and weight affinity assignments follow those in [20]. Furthermore, the multiresolution approach and the corresponding superpixel granularities are exactly the same as in [20] in order to have a fair comparison with this method. The entries of the potential vector $v$ corresponding to the interior superpixels are set to zero, while the ones corresponding to boundary superpixels were set to a fixed value, empirically to one tenth of the maximum entry on the diagonal of the Laplacian matrix. We will show the robustness of the proposed method to this parameter later in this section.

### B. Datasets

MSRA-10k dataset [11] is a superset of the popular ASD [31] dataset and contains 10k images. In each image, there is usually one large salient object with a high contrast to its surrounding. SED2 datasets [35] contain 100 images. Each image contains two salient objects. The saliency detection task in this dataset is relatively simple; however, the challenge is with the objects touching the boundaries. DUTOMRON dataset [14] is one of the largest datasets containing 5168 very challenging images. 25 participants were involved in selecting the images from a wider set of images containing 140k images, and manually labeling the ground truth. ECSSD dataset [12] contains 1000 images with semantically meaningful objects appearing in cluttered backgrounds. Finally, JUDD dataset [42] contains 900 images and is the most challenging database with multiple objects and very high background clutter.

*C. Evaluation Metrics*

Precision is the ratio of the total number of true positives to the total number of detected samples. Recall is the ratio of the number of true positives to the number of true samples. For salient object detection, a true sample is a pixel in the ground-truth segmentation mask $G$. A detected sample at a threshold $\tau$ is the number of nonzero elements in a saliency map thresholded with $\tau$, denoted by $S_\tau$.

$$pre(\tau) = \frac{|G \cap S_\tau|}{|S_\tau|}, \qquad rec(\tau) = \frac{|G \cap S_\tau|}{|G|}$$

(23)

F-measure is defined according to precision and recall values as follows:

$$F_\beta = (1 + \beta^2) \frac{pre(\tau)rec(\tau)}{\beta^2 pre(\tau) + rec(\tau)}$$

(24)

Although $\beta = 1$ is an intuitive selection for equal weighting of precision and recall values, in the salient object detection literature it is common to set $\beta^2 = 0.3$ in order to emphasize more on precision [37]. The reason behind this choice is the fact that it is relatively easier to achieve high recall than high precision. Therefore, in this work, we follow the recommendation in [37].

The receiver operating characteristics (ROC) is the curve plotting true positive rates as a function of false positive rates, denoted as $tpr$ and $fpr$, respectively. The true positive rate is the same as Recall in (23) and the false positive rate is defined as the ratio of the total number of samples that are falsely classified as negative to the total number of negative samples:

$$fpr(\tau) = \frac{|\bar{G} \cap S_\tau|}{|\bar{G} \cap S_\tau| + |\bar{G} \cap \bar{S_\tau}|}$$

(25)

The area under the receiver operating characteristics curve, referred to as AUC-measure, gives a single scalar measure of performance, unlike ROC values at each threshold $\tau$, and hence it is a more general measure of performance.

Precision-recall and ROC curves do not emphasize much on the smoothness within detected salient object regions. For example for a given threshold $\tau$, if every pixel's saliency in the object region is larger than $\tau$, and every background region is smaller than $\tau$, this will yield perfect precision, recall and false positive rate. However, the saliency measures inside the object region can vary widely. In order to take this measure into account, the mean squared error is also used as follows:

$$MSE = \frac{1}{|I|} \sum_{i \in I} (G(i) - S(i))^2$$

(26)

*D. Comparison with the state-of-the-art*

We compare the proposed PSE method with the top-5 performing unsupervised techniques, according to a recent study [43], namely QCUT [20], RBD [15], ST [44], DSR [10] and MC [16]. DRFI [45] was also ranked as one of the top methods in the study; however, it is a supervised method and, hence, it is not included in our comparisons. The saliency maps for the above algorithms were downloaded from the website [30] related to the study in [43]. We further include comparisons with a recently proposed method GP [39]. The saliency maps for GP were extracted using the code provided by the authors.

In Fig. 2, we provide precision-recall curves for the state-of-the-art methods in the datasets mentioned above. One may quickly observe that although some methods such as ST perform well in relatively simpler datasets, such as SED2 and MSRA10k, they fail to provide a good performance in more complex datasets, such as JUDD and DutOmron. One of the top performing methods QCut tends to produce saliency maps that have relatively low precision on higher recalls when compared to other methods. This is evident from the quick *fall* of the curve in this region. Furthermore, due to its hard boundary background assumption in SED2 dataset, this method fails to preserve its leading performance. It is worth noting that the proposed PSE method is consistently one of the top two performing methods in precision-recall curves. PSE does not suffer from the sharp drop in precision at high recall as QCUT. Furthermore, it handles objects touching the image boundary as observed from the plots of SED2 dataset. In Table 1, we provide maximum $F_\beta$ measures for $\beta^2 = 0.3$. In three of the datasets (i.e. MSRA10k, ECSSD and SED2) PSE gives the top performance and on DutOmron and JUDD datasets, it achieves second best performance with a relatively small gap with the leading method.

**Table 1** Maximum $F_\beta$ measures of the proposed PSE and the state-of-the art methods. Top, second and third performing methods are indicated by red, blue and green colors respectively.

|          | PSE    | QCUT   | GP     | RBD    | ST     | DSR    | MC     |
|----------|--------|--------|--------|--------|--------|--------|--------|
| MSRA10k  | 0.8797 | 0.8720 | 0.8610 | 0.8559 | 0.8669 | 0.8346 | 0.8476 |
| DutOmron | 0.6705 | 0.6821 | 0.5838 | 0.6305 | 0.6296 | 0.6265 | 0.6274 |
| ECSSD    | 0.7825 | 0.7778 | 0.7457 | 0.7181 | 0.7504 | 0.7371 | 0.7419 |
| JUDD     | 0.5034 | 0.5093 | 0.4519 | 0.4590 | 0.4571 | 0.4558 | 0.4623 |
| SED2     | 0.8444 | 0.8233 | 0.7704 | 0.8302 | 0.8149 | 0.7900 | 0.7713 |

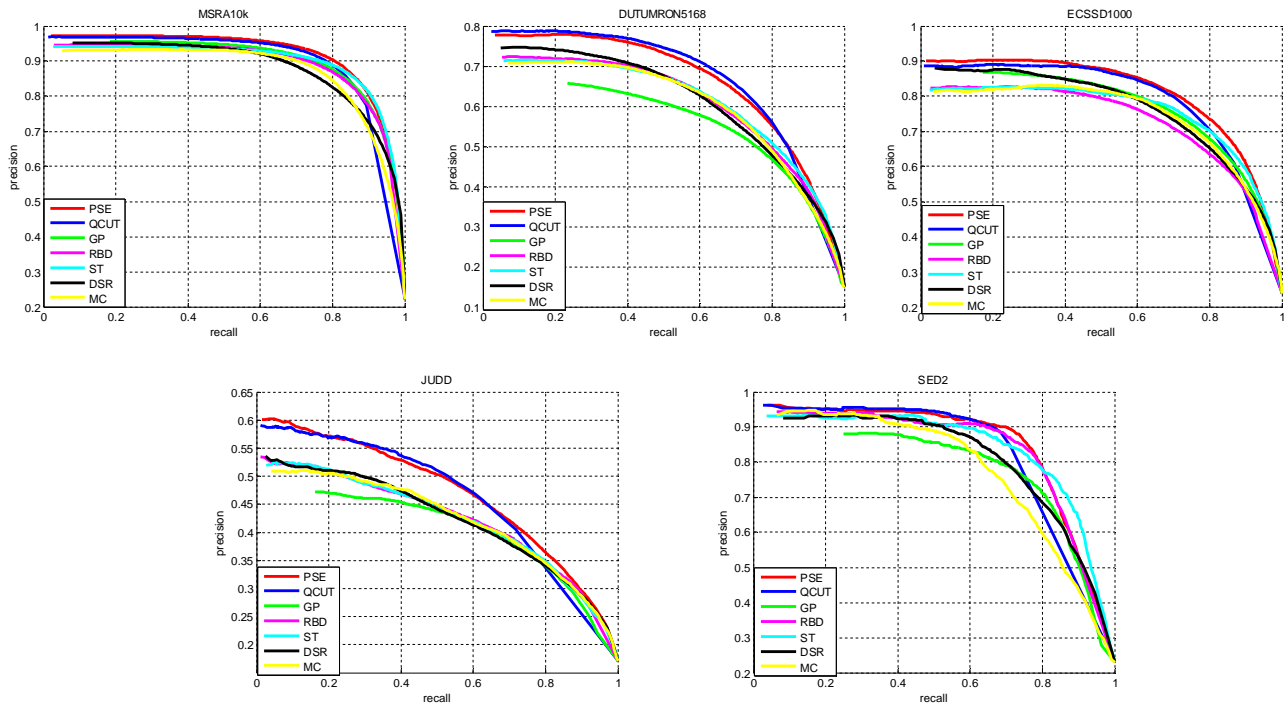

**Fig. 2** Precision-Recall curves for the proposed method PSE, QCUT [20], RBD [15], ST [44], DSR [10], MC [16] and GP [39]. (Top row, from left to right) Curves for datasets MSRA10k, DUTOMRON, ECSSD. (Bottom row, from left to right) Curves for datasets JUDD and SED2.

Since their evaluation is based on different criteria, it is highly likely that one may observe quite different leaderboard in ROC curves and AUC measures than precision-recall curves and F-measures. A good method should perform well with respect to most if not all evaluation metrics. Fig. 3 provides ROC curves for the proposed method against the competing ones. The proposed method PSE is the top performing method in terms of ROC curves, except for SED2 dataset where PSE gives very high recall performance in low false positive rates. Note that one of the leading methods, in precision-recall curves, QCUT does not perform well with respect to ROC curves. The area under the curve (AUC) measures are provided in Table 2.

**Table 2** Performance of the compared methods with respect to AUC measures. Top, second and third performing methods are highlighted in red, blue and green colors, respectively.

|          | PSE    | QCUT   | GP     | RBD    | ST     | DSR    | MC     |
|----------|--------|--------|--------|--------|--------|--------|--------|
| MSRA10k  | 0.9581 | 0.9240 | 0.9579 | 0.9489 | 0.9557 | 0.9528 | 0.9451 |
| DutOmron | 0.9010 | 0.8664 | 0.8672 | 0.8895 | 0.8913 | 0.8945 | 0.8827 |
| ECSSD    | 0.9205 | 0.8836 | 0.9109 | 0.8896 | 0.9092 | 0.9089 | 0.9051 |
| JUDD     | 0.8344 | 0.7740 | 0.7889 | 0.8177 | 0.7964 | 0.8186 | 0.8160 |
| SED2     | 0.8925 | 0.8269 | 0.8717 | 0.8695 | 0.8956 | 0.8922 | 0.8549 |

From Table 2, we observe that PSE is leading in all datasets except for SED2 where it trails the leading method ST by a small margin. GP, DSR and ST also perform well in AUC measures (mostly 2nd and 3rd places); however, they are not as successful in precision-recall curves and F-measure.

Finally, we compare the performance of the compared methods in terms of mean squared errors in Table 3. PSE is the top performing method in ECSSD, JUDD and SED2 databases and the second in MSRA10k and DutOmron. We can conclude that PSE consistently results in leading performance across a wide range of evaluation measures and it is always within the top two.

**Table 3** Performance of the compared methods in terms of MSE errors. Top, second and third performing methods are highlighted in red, blue and green colors, respectively.

|         | **PSE** | QCUT | **GP** | **RBD** | **ST** | **DSR** | MC |
|---------|---------|------|--------|---------|--------|---------|--------|
| MSRA10k | 0.0588  | 0.1064 | 0.0606 | 0.0647 | 0.0568 | 0.0779 | 0.0690 |
| DutOmron | 0.0843 | 0.0869 | 0.1197 | 0.0833 | 0.0911 | 0.0848 | 0.0905 |
| ECSSD   | 0.0948  | 0.1488 | 0.1056 | 0.1156 | 0.0984 | 0.1188 | 0.1067 |
| JUDD    | 0.1264  | 0.1356 | 0.1579 | 0.1484 | 0.1370 | 0.1347 | 0.1296 |
| SED2    | 0.0923  | 0.1456 | 0.1146 | 0.1047 | 0.0929 | 0.1144 | 0.1147 |

For a clearer comparison of the competing methods, we have evaluated the relative percentage performance gain of PSE over the other methods. The gains shown in Table 4 are calculated for each dataset separately and averaged after. Positive gains are favorable for maximum $F_\beta$ and AUC, whereas negatives are favorable for MSE, since it is an error-based measure. It can be observed that the PSE is favorable over all methods in all performance measures on average. It might be also observed that in very few cases PSE suggests a very small improvement. For example, the improvement over QCut in maximum $F_\beta$ is only 0.24%, however in MSE it is %25.48. Since a salient object detection method is best evaluated under taking all the evaluation criteria into account, a general outlook clearly indicates the significance of the relative improvement of PSE over other methods.

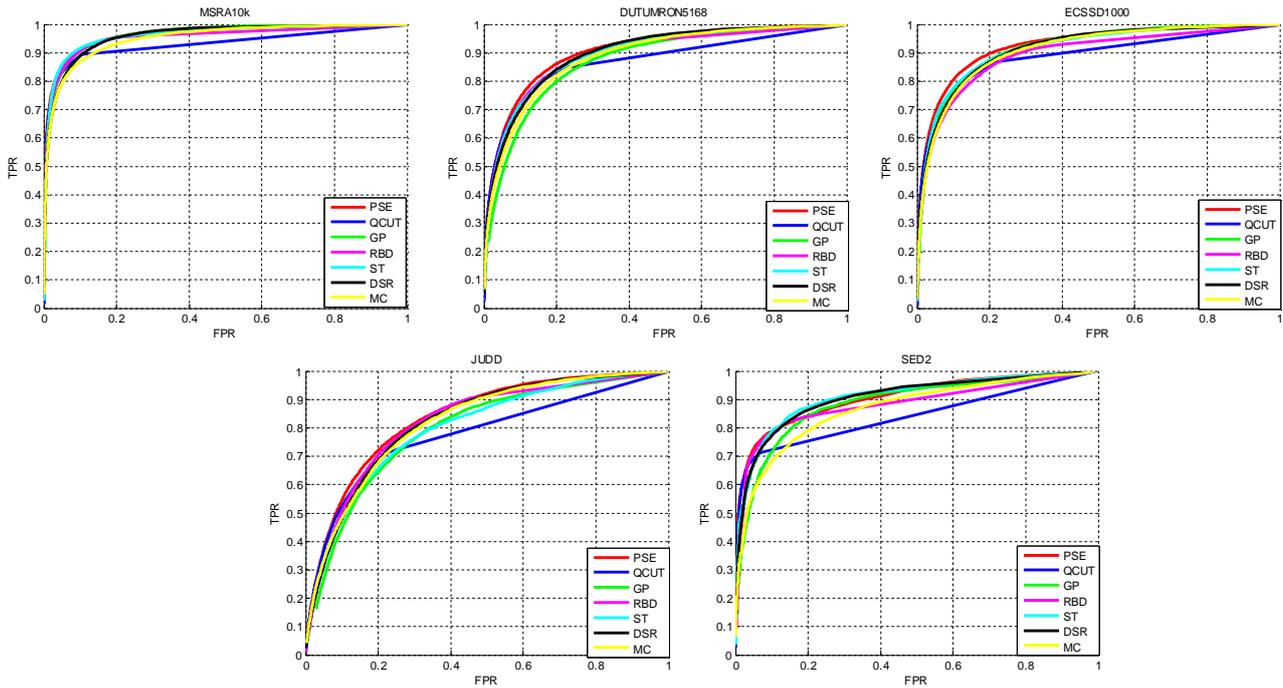

**Fig. 3** ROC curves for the proposed method PSE, QCUT [20], RBD [15], ST [44], DSR [10], MC [16] and GP [39]. (Top row, from left to right) Curves for datasets MSRA10k, DUTOMRON, ECSSD. (Bottom row, from left to right) Curves for datasets JUDD and SED2.

**Table 4** Relative percentage gain of PSE with respect to the state-of-the art methods in all performance measures averaged over all datasets. The preferable performance of PSE is indicated in blue.

|           | QCUT    | GP      | RBD     | ST      | DSR     | MC       |
|-----------|---------|---------|---------|---------|---------|----------|
| Max $F_\beta$ | 0.2382  | 8.5919  | 5.8953  | 5.1999  | 7.1830  | 6.8994   |
| AUC       | 5.5193  | 2.6252  | 2.0846  | 1.4015  | 0.9045  | 2.3606   |
| MSE       | -25.482 | -16.436 | -10.516 | -3.1969 | -14.158 | -10.9569 |

Finally, the average run times for the compared methods are provided in Table 5. Run times denoted by asterisk are calculated on a PC with Intel i7-3740QM CPU @2.70 GHz. Since we do not have access to some of the methods, other run times are copied from the study in [43].

We can see that the run time in our PC is slightly less than the one reported in [43] for QCUT. Hence, it is safe to compare the results accordingly. From Table 5, we can observe that PSE is comparable to the fastest methods and much faster than ST and DSR. Given the leading performance of PSE, slight computational burden over the fastest methods does not prevent favoring our method over others.

**Table 5** Average run times of competing methods.

| | $PSE^*$ | $QCUT^*$ | $QCUT$ | $GP^*$ | RBD | ST | DSR | MC |
|---|---|---|---|---|---|---|---|---|
| Time (sec.) | 1.51 | 1.73 | 1.82 | 1.13 | 0.279 | 79.1 | 10.2 | 0.195 |

*E. Robustness with respect to Boundary Prior*

Next, we will investigate empirically the robustness performance of PSE against variations in the initial boundary potential assignment. As previously noted, we have set a unary potential value on the boundaries of the image to a value proportional to the maximum entry in the diagonal of the Laplacian matrix. Let us denote this value with q.

In Fig. 4, we observe the changes on the precision and recall curves and the receiver operating curves as q changes over a wide range (0,001-10). We can safely say that precision-recall curves are more robust to variations in q than ROC curves. For q values greater than 0.01, one can still observe that ROC curves are also quite robust to this parameter. We have conducted tests on two very different datasets, i.e. ECSSD and SED2. As stated in [43], ECSSD is a dataset where objects are closest to the center and least touching the boundaries. On the other hand, SED2 is a dataset where objects are farthest from the center and most touching the boundaries. Indeed, we observe slightly better precision-recall performance for higher $q$ for ECSSD and lower $q$ for SED2. However, for both datasets, slightly higher values of $q$ are preferable for ROC curves. Higher $q$ values perform better in terms of precision since boundaries are strongly suppressed; however, recall may suffer if the object is touching the boundary. On the other hand, lower $q$ values may lower the precision, but will improve the recall since PSE in this case does not rely too much on the boundary prior. In our experiments, we observed that it is favorable to choose $q = 0.1$ as a moderate value to cover both cases. This value is used in all experiments and all datasets, although better performance than reported in this study may result from tuning this parameter for each dataset. To see the justification more clearly, we also provide the MSE, AUC and maximum $F_\beta$ ($\beta^2 = 0.3$) measures for all $q$ values for ECSSD dataset in Table 6 and for SED2 dataset in Table 7. For the former dataset, the performances for all measures increase as $q$ increases; however, the improvements are not that significant beyond $q = 0.1$. For the latter dataset, we observe a reduced performance for MSE as q increases; however, for AUC and $F_\beta$ measure, the best performance results from moderate values of $q$.

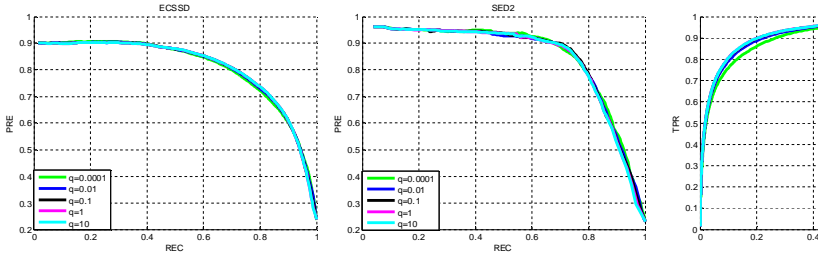

**Fig. 4** (From left to right) PSE performance with respect to Precision-recall and receiver operating curves over ECSSD and SED2 datasets for $q = 0.001, 0.01, 0.1, 1$ and $10$.

**Table 6** MSE, AUC and maximum $F_\beta$ measures for $q = 0.001, 0.01, 0.1, 1$ and $10$ in ECSSD dataset.

| q | 0.001 | 0.01 | 0.1 | 1 | 10 |
|---|---|---|---|---|---|
| MSE | 0.1137 | 0.0976 | 0.0948 | 0.0945 | 0.0945 |
| AUC | 0.9075 | 0.9174 | 0.9205 | 0.9205 | 0.9205 |
| $F_\beta$ | 0.7781 | 0.7815 | 0.7825 | 0.7828 | 0.7829 |

**Table 7** MSE, AUC and maximum $F_\beta$ measures for $q = 0.001, 0.01, 0.1, 1$ and $10$ in SED2 dataset.

| q | 0.001 | 0.01 | 0.1 | 1 | 10 |
|---|---|---|---|---|---|
| MSE | 0.0836 | 0.0886 | 0.0929 | 0.0954 | 0.0955 |
| AUC | 0.8724 | 0.8828 | 0.8925 | 0.8928 | 0.8920 |
| $F_\beta$ | 0.7218 | 0.7347 | 0.7411 | 0.7408 | 0.7383 |

*F. A More Detailed Comparison with GP*

As previously explained, PSE can also be interpreted as a diffusion problem, where a special diffusion matrix **H** is used and the seed vector contains entries that are constant and equal to 1 for each region. Therefore, besides the extensive performance evaluation given above, we provide a more detailed comparison with the most recent diffusion method GP [39]. First, we analyze the effect of the affinity matrix on the performance of GP using our affinity matrix. Next we investigate the effectiveness of the seed vector used in [39] compared to a constant **1** seed vector. We only provide precision-recall curves, since the performance with respect to other metrics are quite similar. As we can observe from **Fig. 5**, GP is heavily dependent on the affinity matrix that is used, as we observe a very sharp performance fall when using our affinity matrix. On the other hand, a minor difference is noticed in performance when using the seed vector **s** and a seed vector of **1**. According to our observations, these effects are mainly due to the graph structure used in [39], where each boundary region is connected to all other boundary regions, making it difficult to assign a boundary region as a salient region. Therefore, assigning zero values to the seed vector to suppress boundary regions is not effective. On the other hand, other graph structures, such as the one used in the proposed method, result in total failure of GP even when using the pre-defined seed vector **s**.

## IV. CONCLUSION

In this paper, we proposed a novel unsupervised salient object detection method based on a probabilistic framework. This method was shown to have several interpretations, such as spectral foreground detection, diffusion-based salient object detection and one-class classification. Such a general framework enjoys a global optimal solution of a criterion that integrates boundary connectivity saliency cue and smoothness constraints. Moreover, based on its interpretations, we showed that it also provides an approximation of the global optimum of another criterion that inherently exploits other saliency cues: local/global contrast and large area cue. Extensive experimental evaluations demonstrated that the proposed method can reach leading performance among the state-of-the-art unsupervised salient object detection methods in a wide variety of datasets with several evaluation criteria.

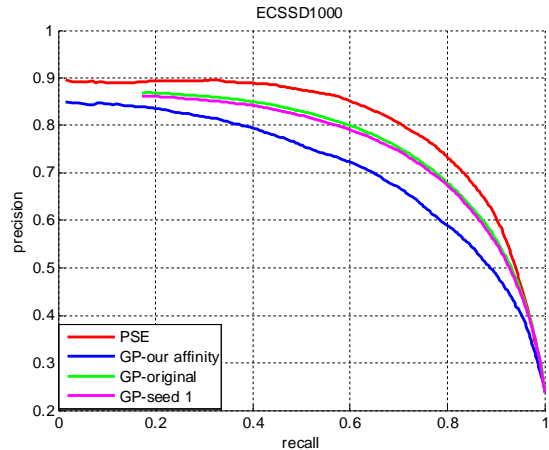

**Fig. 5** Precision-recall curves for ECSSD dataset for PSE, GP, GP with our affinity matrix and GP with seed vector of **1**.

## APPENDIX A

### PROBABILISTIC ANALYSIS OF DIFFUSION METHODS

Here, we provide a probabilistic interpretation of diffusion-based salient object detection methods. Based on this interpretation, we explain why it is expected that PSE is able to produce better saliency maps, when compared to such methods.

Similar to the approach in section II, first let us define the prior probability of selecting a state $x_i$ as the salient region $P(\mathbf{x} = x_i)$, and an *observation random variable* **s** that $P(\mathbf{x} = x_i)$ is dependent on. An example of this observation can be another event of selecting a salient region which can be, for example, another saliency detection method. The $l_1$-normalized saliency result of this supporting model can be denoted as $P(\mathbf{s} = x_i)$. The sample space of both random variables $\mathbf{x}$ and $\mathbf{s}$ is the same and equal to the pixels/regions in the image. This is why samples for both events are defined over the image regions $x_i$. Then, by Bayes theorem, one can write the following:

$$P(\mathbf{x} = x_i | \mathbf{s} = x_j) = \frac{P(\mathbf{s} = x_j | \mathbf{x} = x_i) P(\mathbf{x} = x_i)}{P(\mathbf{s} = x_j)}. \qquad (27)$$

The conditional probability $P(\mathbf{x} = x_i | \mathbf{s} = x_j)$ is most informative when $i = j$, since this means that the probability of selecting image region $x_i$ as a salient one given that the outcome of another salient region selection procedure also selected region $x_i$ as the salient region. We represent $P(\mathbf{x} = x_i | \mathbf{s} = x_i)$ as the $i^{th}$ element of the vector $\mathbf{p_s}$, and $P(\mathbf{s} = x_i)$ as the $i^{th}$ element of vector $\mathbf{s}$. Based on these, we can formulate a new optimization problem, where the goal is to find an optimal $\mathbf{p_s}$ satisfying a similar minimization to that in (1), as follows:

$$\underset{\mathbf{p_s}}{argmin} \left( \sum_{i,j} (p_s(i) - p_s(j))^2 w_{i,j} \right), \qquad (28)$$

$$s.t. \quad p_s^T s = d, d \leq 1.$$

The difference between (28) and (1) is two-fold. First, (28) does not include the background penalization term and second, the constraints are different. It is straightforward to show that $p_s^T s \leq 1$, since $\mathbf{p_s} = P(\mathbf{x} = x_i | \mathbf{s} = x_i)$ corresponds to some portion of the conditional probability. Moreover, the global minimum of (28) cannot be directly obtained as $(D - W)^{-1}s$, since $D - W$ is singular. However, one can find an approximate solution by adding a small noise term to the diagonal of the Laplacian as follows:

$$p_s^* = (D - W + \epsilon I)^{-1}s. \qquad (29)$$

The result in (29) is a regularized solution of (28) and is similar to the approach in [14] as claimed by [39].

The above analysis gives us the formulation of a diffusion based method, where the diffusion matrix is the inverse Laplacian matrix. An important observation related to the Bayesian interpretation of diffusion-based methods is that these methods are interested in the conditional probability $\mathbf{p_s} = P(\mathbf{x} = x_i | \mathbf{s} = x_i)$ and discard information about $\mathbf{p_s} = P(\mathbf{x} = x_i | \mathbf{s} = x_j)$ when $i \neq j$. Based on this, we can conclude that this result is dependent on the observation $\mathbf{s}$ that is interpreted as another salient region selection event. On the other hand, by the formulation of Eq. (1), PSE does not include such dependency to a pre-estimated saliency map, it rather suppresses confident background regions. This is a more trustable approach since background initialization is much easier than foreground estimation. Furthermore, approaches such as selecting the seed vector as 0 on the image boundaries and 1 in the foreground to use boundary connectivity cue results in noisy saliency maps, since all interior regions are given the same initial saliency confidence.

Above, we have presented a probabilistic interpretation for a specific diffusion based method [14]. We can give this interpretation for other diffusion based methods as well. For example let us consider the work presented in [40], where a manifold ranking method was proposed that exploits the manifold of the given graph structure. In [41], it was shown that this manifold ranking corresponds to the following unconstrained minimization:

$$\underset{f}{argmin} \frac{1}{2}\left( \mu \sum_i (f_i - s_i)^2 + \sum_{i,j} \left( \frac{f_i}{\sqrt{d_{ii}}} - \frac{f_j}{\sqrt{d_{jj}}} \right)^2 w_{i,j} \right). \tag{30}$$

In (30), $f$ is the result of the ranking query and $s$ is the initial confidence. The optimal solution is obtained by:

$$f^* = \frac{\mu}{1+\mu} \left( I - \frac{D^{-1/2} W D^{-1/2}}{1+\mu} \right)^{-1} s. \tag{31}$$

Graph based manifold ranking [14], which is yet another diffusion based method, can also be formulated in our Bayesian interpretation as follows:

$$\underset{p_s}{argmin} \frac{1}{2}\left( \sum_{i,j} \left( \frac{p_s(i)}{\sqrt{d_{ii}}} - \frac{p_s(j)}{\sqrt{d_{jj}}} \right)^2 w_{i,j} + \mu \sum_i (p_s(i))^2 \right),$$

$$s.t. \quad p_s^T s = d, \qquad d \le 1. \tag{32}$$

The global optimal solution $p_s^*$ is exactly the same as $f^*$ except for the normalization term $\frac{\mu}{1+\mu}$, instead we have a term related to both $\mu$ and $s$ which is ensured by the constraint $p_s^T s = d$. Indeed, (30) is a special case of (32), where we restrict the joint probability of $x$ and $s$ in the region of interest to ensure the normalization value $\frac{\mu}{1+\mu}$ in (31). One can also show that $d$ and $\mu$ is proportional. A large $\mu$ value also ensures a large $d$ value, which means that the confidence on the observation $s$ is large. In the extreme case, i.e. when $p_s^T s = 1$, events $x$ and $s$ correspond to exactly same random variable.

Comparing (32) and (28), we can observe the difference on the smoothing term. Instead of directly ensuring similar posterior probabilities for samples with similar features as in (28), the effect of similarity enforcement was normalization of posterior probabilities in (32). This means that while enforcing similarity between similar features, the effects of the surrounding of both samples are considered.

There are two shortcomings of this approach in terms of saliency: first the dependency on the parameter $\mu$, second similarity enforcement based on support regions. Selecting an optimal $\mu$ value is an open problem. However, a bigger problem is the effect of similarity enforcement based on support regions instead of the regions themselves. This results in different labeling of parts of regions closer to object edges and the ones in the interior object region.

We have given a Bayesian interpretation of diffusion based methods and explained their shortcomings based on this approach. These examples can be populated by giving other optimization problems resulting in different diffusion matrices. However, due to lack of space we omit giving more examples. As a result of our investigation both experimentally and theoretically, we can state that these methods require a good initialization. Therefore diffusion based salient object methods [16],[14],[39] require a good post-processing technique and special graph designs to produce an accurate saliency map. On the other hand, PSE does not require any foreground confidence nor does it require any refining procedures to obtain an accurate saliency map.

APPENDIX B

PROBABILISTIC INTERPRETATION OF QUANTUM-CUTS

Here we provide a probabilistic interpretation for QCut. Similar to (1), with the restriction $w_{i,j} = w_{j,i}$, (17) can be rewritten as follows:

$$\underset{P(\mathbf{x})}{argmin} \left( \sum_i P(\mathbf{x} = x_i) v_i + \frac{1}{2} \sum_{i,j} \left( \sqrt{P(\mathbf{x} = x_i)} - \sqrt{P(\mathbf{x} = x_j)} \right)^2 w_{i,j} \right),$$

$$s.t. \quad \sum_i P(\mathbf{x} = x_i) = 1. \tag{33}$$

The difference between (1) and (33) is the fact that the former penalizes the distance between estimated probabilities whereas the latter penalizes the distance between the square-root of the probabilities. We have already observed that PSE generates much more accurate saliency maps compared to QCut, especially in terms of high recall in this study. For symmetric affinity matrices and a non-zero prior $v$, we have justified this performance difference by the relation of both methods via diffusion maps [38]. We can also come to the same conclusion observing that (33) may produce much sharper saliency maps compared to (1). Since the probability values are smaller than (or equal to) 1, minimizing the distance between the square-root of the probabilities would result in significant changes in saliency maps between the regions with small contrast. However, (1) results into smoother regions, ensuring a higher recall.

APPENDIX C

DIFFUSION METHODS AS CLASSIFICATION METHODS

As has been described in Subsection II.D, PSE can be interpreted as a one-class classification problem, trying to model the image at hand. Based on our observation in Subsection II.A related to the similarity of PSE and diffusion methods, we can also cast such methods as classification methods. However, since they usually exploit a binary seed vector $s$, which might be the output of another saliency detection method, they solve a two-class problem. In the case where the exploited seed vector $s$ contains real values, diffusion methods solve a regression model trying to map the image region's features to the values of $s$. In other words, they are trying to find a projection vector that optimally regresses the image features to the seed vector, which might be an output of another saliency detection method or some heuristics based initialization. In both cases, it is clear that their output $y^*$ will be dependent on $s$.

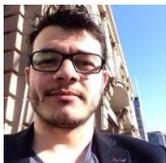 **Çağlar Aytekin** was born in Turkey in 1987. He received the B.Sc. degree from the Electrical and Electronics Engineering Department from Middle East Technical University, Ankara Turkey in 2008. He received the M.Sc. department with Signal Processing major from Middle East Technical University, Ankara, Turkey in 2011. He is currently a Ph.D student and researcher in Department of Signal Processing, Tampere University of Technology. He has published 16 papers in international conferences and journals. He has won the the ICPR 2014 – Best Student Paper Award for his work in unsupervised salient object detection. His research interests are, visual saliency, semantic segmentation, quantum mechanics based computer vision methods and deep learning.

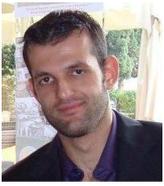

**Alexandros Iosifidis** (SM'16) received the Diploma (5-year degree) and Master of Engineering in Electrical & Computer Engineering from Democritus University of Thrace in 2008 and 2010, respectively. He also received a Ph.D. in Informatics from Aristotle University of Thessaloniki in 2014. Currently, he is a postdoctoral researcher at the Multimedia Research Group of the Department of Signal Processing in Tampere University of Technology, holding the prestigious Academy Postdoctoral Research Fellowship. In 2015, he was awarded a two-year postdoc fellowship given by TUT Foundation to five (in total) postdoctoral researchers in all scientific fields. Dr. Iosifidis is a Senior member of IEEE and member of EURASIP and the Finnish Academy of Science and Letters. He has participated in 6 research projects funded by EU, Greek and Finnish funds. He has (co-)authored 33 journal papers, 56 conference papers and 4 book chapters. His research interests are in the areas of pattern recognition and machine learning, with applications mainly in images and videos.

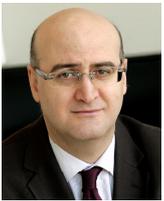

**Moncef Gabbouj** received his BS degree in electrical engineering in 1985 from Oklahoma State University, Stillwater, and his MS and PhD degrees in electrical engineering from Purdue University, West Lafayette, Indiana, in 1986 and 1989, respectively. Dr. Gabbouj is a Professor of Signal Processing at the Department of Signal Processing, Tampere University of Technology, Tampere, Finland. He was Academy of Finland Professor during 2011-2015. He held several visiting professorships at different universities. His research interests include multimedia content-based analysis, indexing and retrieval, machine learning, nonlinear signal and image processing and analysis, voice conversion, and video processing and coding. Dr. Gabbouj is a Fellow of the IEEE and member of the Academia Europaea and the Finnish Academy of Science and Letters. He is the past Chairman of the IEEE CAS TC on DSP and committee member of the IEEE Fourier Award for Signal Processing. He served as Distinguished Lecturer for the IEEE CASS. He served as associate editor and guest editor of many IEEE, and international journals. Dr. Gabbouj was the recipient of the 2015 TUT Foundation Grand Award, the 2012 Nokia Foundation Visiting Professor Award, the 2005 Nokia Foundation Recognition Award, and several Best Paper Awards. He published over 650 publications and supervised 40 doctoral theses.